\documentclass[conference]{IEEEtran}
\IEEEoverridecommandlockouts

\usepackage{cite}
\usepackage{amsmath,amssymb,amsfonts}
\usepackage{algorithmic}
\usepackage{graphicx}
\usepackage{textcomp}
\usepackage{xcolor}
\usepackage{tikz}
\usepackage{url}
\usetikzlibrary{positioning, arrows.meta, calc,decorations.pathreplacing}
\usepackage{booktabs}   
\usepackage{siunitx}    

\def\BibTeX{{\rm B\kern-.05em{\sc i\kern-.025em b}\kern-.08em
    T\kern-.1667em\lower.7ex\hbox{E}\kern-.125emX}}

\begin{document}

\title{Non-Destructive Quantification of Urea Adulteration in Bovine Milk Using Transmittance Multispectral Imaging}

\author{
\IEEEauthorblockN{Sharukshan Niranjan}
\IEEEauthorblockA{\textit{Dept. of Electrical and}\\
\textit{Electronic Eng.}\\
\textit{University of Peradeniya}\\
Peradeniya, Sri Lanka\\
n.sharukshan@gmail.com}
\and
\IEEEauthorblockN{Iresha Ranaweera}
\IEEEauthorblockA{\textit{Dept. of Electrical and}\\
\textit{Electronic Eng.}\\
\textit{University of Peradeniya}\\
Peradeniya, Sri Lanka\\
ranaweeraagil@gmail.com}
\and
\IEEEauthorblockN{Tharindu Chandrarathne}
\IEEEauthorblockA{\textit{Dept. of Electrical and}\\
\textit{Electronic Eng.}\\
\textit{University of Peradeniya}\\
Peradeniya, Sri Lanka\\
thar1994@gmail.com}
\and
\IEEEauthorblockN{Kalana Dissanayaka}
\IEEEauthorblockA{\textit{Dept. of Food Science and}\\
\textit{Technology}\\
\textit{University of Peradeniya}\\
Peradeniya, Sri Lanka\\
kalanad@agri.pdn.ac.lk}
\and
\IEEEauthorblockN{Roshan Godaliyadda}
\IEEEauthorblockA{\textit{Dept. of Electrical and}\\
\textit{Electronic Eng.}\\
\textit{University of Peradeniya}\\
Peradeniya, Sri Lanka\\
roshang@eng.pdn.ac.lk}
\and
\IEEEauthorblockN{Vijitha Herath}
\IEEEauthorblockA{\textit{Dept. of Electrical and}\\
\textit{Electronic Eng.}\\
\textit{University of Peradeniya}\\
Peradeniya, Sri Lanka\\
vijitha@eng.pdn.ac.lk}
\and
\IEEEauthorblockN{Parakrama Ekanayake}
\IEEEauthorblockA{\textit{Dept. of Electrical and}\\
\textit{Electronic Eng.}\\
\textit{University of Peradeniya}\\
Peradeniya, Sri Lanka\\
mpb.ekanayake@ee.pdn.ac.lk}
\and
\IEEEauthorblockN{Janak Vidanarachchi}
\IEEEauthorblockA{\textit{Dept. of Animal Science}\\
\textit{University of Peradeniya}\\
Peradeniya, Sri Lanka\\
janakvid@pdn.ac.lk}
}

\maketitle


\begin{abstract}
Adulteration of bovine milk using urea remains a major food quality and health concern, motivating the development of rapid and quantitative screening tools. Conventional approaches, including laboratory-based analytical methods and spectroscopic techniques, have been used for urea detection; however, many remain less suitable for rapid, low-cost routine screening due to requirements such as specialized instrumentation, sample preparation, chemical reagents, or laboratory operation. This study introduces a pragmatic, cost-effective, accurate, and laboratory-validated MSI-based method for quantitative urea estimation under controlled density conditions using a multispectral-imaging-based regression framework. An in-house-built multispectral imaging system operating in twelve discrete spectral bands (365--940~nm) was used to acquire multispectral images of milk samples prepared with controlled urea addition and water for density balancing. Fresh milk was obtained on the day of image acquisition, and the specific gravity of the milk was verified to be 1.032 at \SI{20}{\celsius} using a hydrometer. Multiple linear regression provided an initial mapping with a high validation $R^2$ of 0.9599, while a feed-forward neural network further improved predictive performance with a validation $R^2$ of 0.9773. These results demonstrate the feasibility of transmittance multispectral imaging for accurate, non-destructive urea quantification under controlled density-balanced conditions, supporting its potential as a rapid screening approach for milk-quality assessment.
\end{abstract}


\section{Introduction}
\label{sec:introduction}

Bovine milk is a socially and economically vital dietary staple, globally recognized for its highly nutritive profile \cite{b1,b2}. Milk is a complex biological fluid composed primarily of water, proteins, lipids, minerals, and vitamins, and plays a crucial role in nutrition and immune support~\cite{b3,b4}. Being easily digestible, milk is crucial for the nutritional requirements of vulnerable groups like infants and the elderly \cite{b5,b2}. Driven by this universal demand, global milk production surpassed 950 million tons in 2024 \cite{b3}.

However, the existing gap between high demand and limited supply, combined with profit motives, has fueled widespread intentional milk adulteration \cite{b4,b6}. Remaining among the most counterfeited foods worldwide, milk accounted for 14\% of all reported food fraud incidents between 1980 and 2010 \cite{b7}. Fraudulent practices typically involve adding inexpensive or dangerous adulterants such as water, urea, formalin, and melamine to increase volume or mask inferior quality \cite{b6,b8}. Consumption of adulterated milk triggers a severe public health crisis, linked to gastrointestinal disorders, cardiovascular diseases, and cancer \cite{b6,b3}. This severity is underscored by the 117, 149 cancer-related deaths in Pakistan in 2020, where contaminated milk posed a significant risk \cite{b8}. Adulteration also causes immense economic damage through market instability and increased regulatory costs \cite{b5,b2}.

Among adulterants, the intentional addition of extraneous nitrogenous compounds like urea is a major food safety concern \cite{b9}. While naturally present at approximately 70 mg/100 mL, fraudulent producers add commercial urea in much higher quantities (90 to 4000 mg per liter) to economically mask water dilution \cite{b1,b6}. Urea falsely elevates the non-protein nitrogen (NPN) concentration, providing the illusion of high protein content and increased consistency to create a toxic ``synthetic milk'' \cite{b2,b9,b5}. Chronic consumption of this adulterant severely strains the kidneys, negatively impacts the gastrointestinal tract, and can lead to indigestion, ulcers, and cancer \cite{b10,b3,b2}. Despite these risks, fraudulent practices persist due to inadequate monitoring and the lack of rapid detection technologies \cite{b4}. Although laboratory-based analytical methods such as liquid chromatography and mass spectrometry provide accurate quantification, they are less suited for routine on-site screening because they require specialized equipment, trained personnel, longer analysis times, and higher operational costs \cite{b8,b10,b6}. Recent spectroscopic, electrochemical, and point-of-care approaches have also shown promise for milk adulteration detection, motivating the investigation of complementary low-cost imaging-based screening methods.

Motivated by these limitations, this study proposes a rapid, non-destructive, and quantitative method for estimating urea adulteration in cow milk using an in-house-built multispectral imaging (MSI) system. The system captures selected spectral bands from 365--940~nm, and regression models are used to map the extracted spectral information to density-balanced urea concentration. The proposed framework demonstrates the feasibility of MSI-based urea estimation under controlled density-balanced laboratory conditions and provides a basis for broader validation under practical dairy supply-chain conditions.

\section{Materials and Methods}
\label{sec:materials}

\subsection{Multispectral Imaging System}
\label{subsec:msi_system}
In contrast to RGB images which contain only three visible bands, multispectral images contain a larger number of spectral bands from near-visible ultraviolet (NUV) to near-infrared (NIR). Thus, this broader spectral coverage allows for more information about the composition of substances to be captured which is particularly useful for food quality assessment. However, commercially available multispectral cameras are capable of a high spectral and spatial resolution but at a significantly higher cost of purchase. Moreover, when only selected wavelength regions are informative for the target application, acquiring a wide range of spectral data may introduce redundant information. Therefore, an economical in-house-built MSI system \cite{msi2024dual} capable of capturing selected wavelengths from the near-ultraviolet (NUV) to the near-infrared (NIR) range was employed for this analysis. This imaging system has already been successfully utilized for soil classification using moisture-induced spectral dynamics~\cite{soil2025} and aflatoxin contamination level estimation in food~\cite{aflatoxin2024}.

\begin{figure}[!t]
\centering
\includegraphics[width=0.9\linewidth]{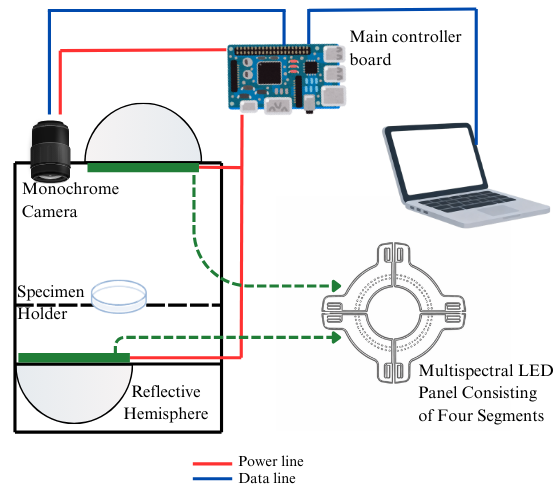}
\caption{Schematic diagram of MSI system used for detecting urea in milk.}
\label{fig:msi_schematic}
\end{figure}

As depicted in Fig. \ref{fig:msi_schematic}, the imaging system is composed of several principal components. The capturing of the images was carried out using a CMOS monochrome vision camera (FLIR BFS-U3-12Y3M, 1.3MP, USB3 Vision v1.0, Resolution - 1280 × 1024, ADC – 10 bit) coupled with an onsemi PYTHON 1300 image sensor. The illumination module consisted of twelve discrete narrow-band LEDs with peak wavelengths of 365nm, 405nm, 473nm, 530nm, 621nm, 660nm, 735nm, 770nm, 830nm, 850nm, 890nm and 940nm. In addition, this also aligns well with the frequency response of the camera, which was 350nm – 1080nm. The image capturing and data transfer were facilitated using an Arduino Duo microcontroller board and a computer.

\subsection{Sample Preparation}
\label{subsec:sample_prep}

In order to evaluate the proposed methods for adulteration analysis, a series of milk samples was prepared using pure bovine milk as the base material. Analytical-grade urea was weighed and added to the milk to obtain ten nominal adulteration levels: 0, 0.1, 0.3, 0.6, 1.0, 1.5, 2.0, 3.0, 4.0, and 5.0 g/100 mL.

Each sample was prepared as a 50 mL mixture. For each concentration level, eight independently prepared replicate samples were used, resulting in a total of 80 physical sample image sets across the ten concentration levels. Following the addition of urea, the density of each sample was adjusted, and the mixture was stirred gently until the urea was fully dissolved and the sample was visually homogeneous. Particular care was taken to minimize foaming and bubble formation, since these could negatively affect the captured images.

Finally, each sample was labeled with its concentration level and replicate ID to ensure traceability throughout image acquisition and subsequent analysis.

\subsection{Density Control}
\label{subsec:density_control}
Since the addition of urea can affect the bulk density of milk, all samples were prepared under controlled density conditions in order to minimize the confounding effect between urea concentration and density related optical variations. In this study, the target density corresponded to a lactometer reading of 32~\cite{jananika2025rawmilk}, which is equivalent to a milk density of 1.032 g/mL and the density of water was taken as 0.998 g/mL.

To preserve this target density while increasing the urea content, an appropriate amount of water was added to each sample. Thus, although the addition of urea increased the overall density of the mixture, the added water compensates for this effect and helps maintain density consistency across the samples. Assuming that the volume contribution from the added urea is negligible, the required water volume for each selected urea mass was estimated using a simple mass-balance relationship:

\begin{equation}
V_w = \frac{m_{\text{urea}}}{\rho_{\text{milk}} - \rho_{\text{water}}}.
\end{equation}

where $V_w$ is the volume of water added, $m_{\text{urea}}$ is the mass of added urea, $\rho_{\text{milk}}$ is the density of milk, and $\rho_{\text{water}}$ is the density of water.

Accordingly, for each concentration level, the required mass of urea was first weighed. Subsequently, the corresponding volume of deionized water was calculated and added during sample preparation. This procedure was applied consistently to all samples hence that variations observed in the multispectral measurements could be attributed mainly to differences in urea concentration rather than unintended density changes.
\begin{figure}[!t]
\centering
\includegraphics[width=1\linewidth]{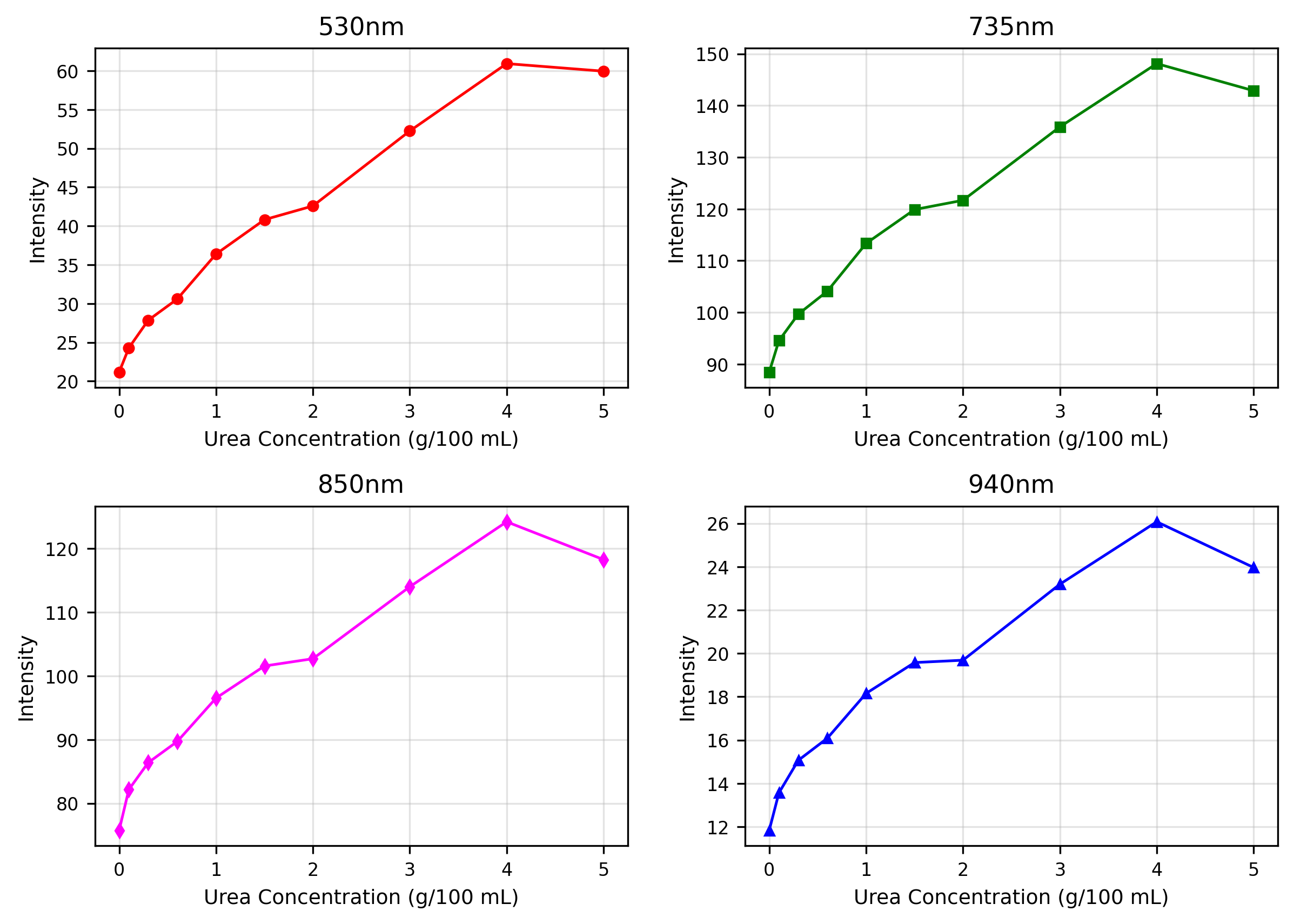}
\caption{Variations of intensities with density balanced urea concentration at wavelengths 530nm, 735nm, 850nm, and 940nm for all measured levels in milk}
\label{fig:wavelength_vs_adulteration}
\end{figure}

\subsection{Multispectral Image Acquisition}
\label{subsec:image_capture}

Prior to image acquisition, the imaging system was calibrated to ensure consistent capture conditions across all samples. The camera focus was first adjusted to obtain clear and well-defined images, the aperture was set to provide adequate sensor illumination without saturation, and the zoom level was adjusted so that the complete specimen was captured while minimizing irrelevant background regions. Following calibration, the system was operated in transmittance mode, and image acquisition was carried out under the controlled density conditions described in Section~\ref{subsec:density_control}. For each prepared milk sample, the sample container was placed at the same position within the imaging system to maintain geometric consistency.

Each sample was sequentially illuminated using the twelve discrete LEDs, producing 12 spectral images corresponding to the selected wavelength bands. In addition, one dark-current image was captured under the same camera settings without LED illumination for subsequent dark-current correction. Accordingly, each multispectral image set consisted of 13 images: 12 illuminated spectral images and one dark-current image. Since 10 adulteration levels with 8 independently prepared replicates per level were used, a total of 80 multispectral image sets were acquired, corresponding to 1040 raw images. Of these, 960 illuminated spectral images were used for spectral feature extraction after dark-current correction.

\subsection{Preprocessing} 
\label{subsec:preprocess} 
Before feature extraction and modeling, the obtained multispectral images were subjected to preprocessing in order to minimize the impact of sensor noise and enhance the quality of spectral data. This is crucial as the captured response could be impacted by sensor offset noise and random noise. Dark current subtraction was initially implemented to minimize the effect of sensor baseline noise, which is caused by thermally produced electrons inside the imaging sensor and could cause a non-zero response even in the absence of illumination \cite{Widenhorn2007}. Dark current correction was achieved by capturing a dark image without the light sources but using the same camera parameters and then is subtracted from all successive illuminated images as shown in the following equation. 
\begin{equation} B[\lambda] = A[\lambda] - D, \end{equation} 
where $A[\lambda]$ refers to the raw image acquired using wavelength $\lambda$, $D$ is the dark current image, and $B[\lambda]$ is the dark current-corrected image. 

Following dark current correction, a region of interest (ROI) was cropped to a fixed size of $100 \times 100$ pixels to isolate the sample and reduce background interference. However, residual random noise may still persist in the corrected images. To mitigate this an adaptive Wiener filter was applied. Unlike fixed smoothing methods, the adaptive Wiener filter adapts to local image statistics, enabling stronger smoothing in uniform regions and preserving edges in areas with higher variance \cite{2021wiener}. Assuming that the variance of the white noise is spatially homogeneous over the entire image the global noise variance was calculated as $\sigma_{n}^{2}$. Finally, the Wiener-filtered image $\hat{B}(x,y,\lambda)$ was obtained as 
\begin{equation} \hat{B}(x,y,\lambda) = \mu_{L} + \frac{\sigma_{B}^{2} - \sigma_{n}^{2}}{\sigma_{B}^{2}} \left[ B(x,y,\lambda) - \mu_{L} \right] \end{equation} 
where $\mu_{L}$ and $\sigma_{B}^{2}$ are the local mean and variance for the adopted $3 \times 3$ neighbourhood. A window size of $3 \times 3$ pixels was considered an optimal trade-off between noise reduction and maintenance of informative image features. A larger window can lead to over-smoothing and suppress subtle image variations, whereas a smaller window may not reduce noise sufficiently.
\begin{figure}[!t]
\centering
\resizebox{\linewidth}{!}{
\begin{tikzpicture}[font=\sffamily\large, xscale=1.3, yscale=1.3]

\definecolor{c1}{RGB}{0,0,130}      
\definecolor{c2}{RGB}{0,0,255}      
\definecolor{c3}{RGB}{0,100,255}    
\definecolor{c4}{RGB}{0,200,255}    
\definecolor{c5}{RGB}{0,255,150}    
\definecolor{c6}{RGB}{0,255,0}      
\definecolor{c7}{RGB}{150,255,0}    
\definecolor{c8}{RGB}{255,255,0}    
\definecolor{c9}{RGB}{255,190,0}    
\definecolor{c10}{RGB}{255,100,0}   
\definecolor{c11}{RGB}{255,0,0}     
\definecolor{c12}{RGB}{150,0,0}     

\begin{scope}[shift={(0,0)}]
    \foreach \i/\col in {12/c12, 11/c11, 10/c10, 9/c9, 8/c8, 7/c7, 6/c6, 5/c5, 4/c4, 3/c3, 2/c2, 1/c1} {
        \pgfmathsetmacro{\offset}{(\i-1)*0.15}
        \begin{scope}[shift={(\offset, \offset)}]
            \fill[\col] (0,0) rectangle (4,4);
            \draw[white, very thin, step=0.1] (0,0) grid (4,4);
            \draw[black, thin] (0,0) rectangle (4,4);
        \end{scope}
    }
    
    \draw[black, thick, fill=c1, fill opacity=0.6] (0, 3.6) rectangle (0.4, 4);
    
    \draw[Stealth-Stealth] (-0.2, 0) -- (-0.2, 4) node[midway, above, rotate=90] {100 Pixels};
    \draw[Stealth-Stealth] (0, -0.2) -- (4, -0.2) node[midway, below] {100 Pixels};
    
    \draw[Stealth-Stealth] (4.15, -0.15) -- (5.8, 1.5) node[midway, sloped, below=4pt] {12 Spectral Bands};
    
    \node at (2, -1) {\LARGE (a)};
\end{scope}

\begin{scope}[shift={(7.5,0)}]
    \foreach \i/\col in {12/c12, 11/c11, 10/c10, 9/c9, 8/c8, 7/c7, 6/c6, 5/c5, 4/c4, 3/c3, 2/c2, 1/c1} {
        \pgfmathsetmacro{\offset}{(\i-1)*0.15}
        \begin{scope}[shift={(\offset, \offset)}]
            \fill[\col] (0,0) rectangle (4,4);
            \draw[white, thin, step=0.4] (0,0) grid (4,4);
            \draw[black, thin] (0,0) rectangle (4,4);
        \end{scope}
    }
    
    \foreach \x [count=\n] in {0.2, 0.6, 1.0, 1.4, 1.8, 2.2, 2.6, 3.0, 3.4, 3.8} {
        \node[text=white] at (\x, 3.8) {\scriptsize \n};
    }
    \foreach \y [count=\n] in {3.4, 3.0, 2.6, 2.2, 1.8, 1.4, 1.0, 0.6, 0.2} {
        \pgfmathsetmacro{\val}{int((\n+1)*10)}
        \node[text=white] at (3.8, \y) {\scriptsize \val};
    }
    
    \draw[Stealth-Stealth] (-0.2, 0) -- (-0.2, 4) node[midway, above, rotate=90] {10 Superpixels};
    \draw[Stealth-Stealth] (0, -0.2) -- (4, -0.2) node[midway, below] {10 Superpixels};
    
    \node at (2, -1) {\LARGE (b)};
\end{scope}

\draw[red, -{Stealth[length=3.5mm, width=2.5mm]}, thick] (0.4, 3.8) -- (7.4, 3.8);

\begin{scope}[shift={(15.0,0)}]
    \foreach \i/\col in {1/c1, 2/c2, 3/c3, 4/c4, 5/c5, 6/c6, 7/c7, 8/c8, 9/c9, 10/c10, 11/c11, 12/c12} {
        \fill[\col] (\i*0.3-0.3, 3.25) rectangle (\i*0.3, 5.65);
        \fill[\col] (\i*0.3-0.3, 0) rectangle (\i*0.3, 2.4);
    }
    
    \draw[white, thin, xstep=0.3, ystep=0.48] (0, 3.25) grid (3.6, 5.65);
    \draw[black, thin] (0, 3.25) rectangle (3.6, 5.65);
    
    \draw[white, thin, xstep=0.3, ystep=0.48] (0, 0) grid (3.6, 2.4);
    \draw[black, thin] (0, 0) rectangle (3.6, 2.4);
    
    \foreach \y in {2.55, 2.69, 2.82, 2.96, 3.10} {
        \fill[black] (1.8, \y) circle (0.8pt);
    }
    
    \foreach \n in {1,2,3,4,5,6,7,8,9,10,11,12} {
        \node[text=black] at (\n*0.3-0.15, 5.9) {\scriptsize \n};
    }
    
    \draw[Stealth-Stealth] (-0.3, 0) -- (-0.3, 5.65) node[midway, above, rotate=90] {100 Superpixels};
    
    \node at (1.8, -1) {\LARGE (c)};
\end{scope}

\end{tikzpicture}
}
\caption{Development of data matrix (a) Data matrix corresponding to 100 × 100 pixel cropped image for a given sample (b) 10×10 superpixel data matrices developed for all spectral bands by subdividing each 100×100 pixel image into 10×10 pixel sections (c) Final matrix with dimensions of 100 × 12 }
\label{fig:spectral_data}
\end{figure}
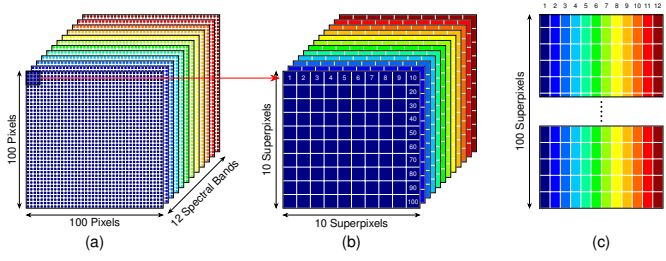

\subsection{Data Representation, Partitioning, and Spectral Signature}
\label{subsec:data_spectral}

Once preprocessing was complete, each $100 \times 100$ pixel ROI was divided into non-overlapping $10 \times 10$ pixel regions. The mean intensity of each region was computed and treated as one superpixel feature. Therefore, 100 superpixel intensities were obtained per spectral feature, as shown in Fig.~\ref{fig:spectral_data}(a) and Fig.~\ref{fig:spectral_data}(b). Since 12 spectral bands were used, each multispectral image set was represented by a $100 \times 12$ feature matrix, where each row corresponded to one superpixel and each column corresponded to one spectral band, as shown in Fig.~\ref{fig:spectral_data}(c).

For model development, seven concentration levels (0.0, 0.1, 0.6, 1.0, 2.0, 3.0, and 5.0 g/100 mL) were used for training and testing, whereas three intermediate concentration levels (0.3, 1.5, and 4.0 g/100 mL) were reserved exclusively for independent validation to assess interpolation to unseen urea concentrations. Each concentration level comprised eight independent replicate image sets, resulting in 56 image sets for model development and 24 image sets for validation. The 56 image sets were partitioned using a stratified 75/25 split at the replicate image-set level, yielding 42 training and 14 testing image sets. Partitioning was performed before superpixel extraction to ensure that all 100 superpixels derived from a given image set remained within the same partition, thereby preventing data leakage between the training and testing datasets.

\begin{figure}[!t] \centering \includegraphics[width=1\linewidth]{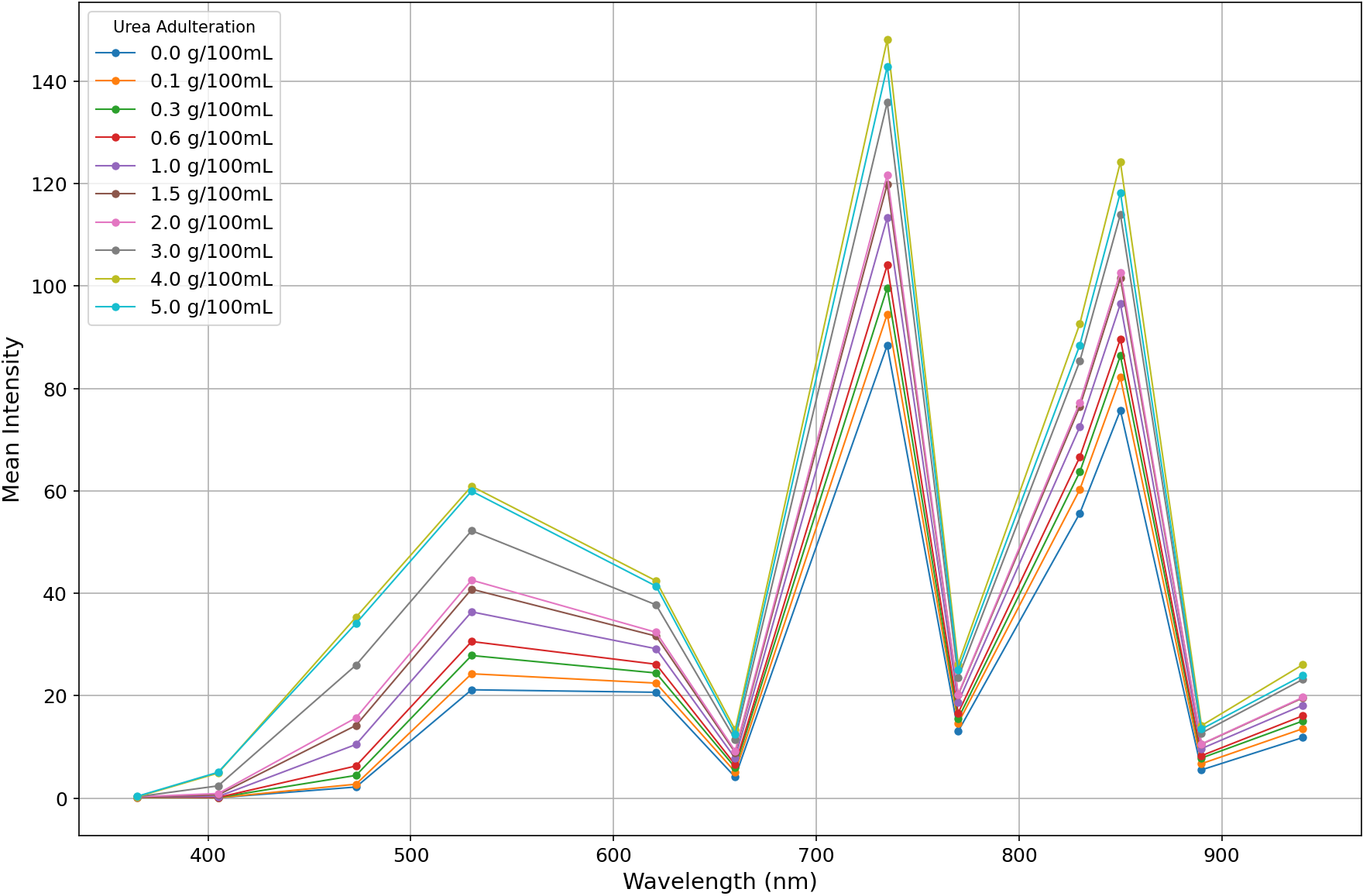} \caption{Mean spectral signatures of milk samples with varying urea concentrations } \label{fig:spectral_sig} \end{figure}

Since each image set yielded 100 superpixel feature vectors of 12 multispectral bands , the final MSI superpixel datasets comprised a $4200 \times 12$ training feature matrix, a $1400 \times 12$ testing feature matrix, and a $2400 \times 12$ validation feature matrix. In addition, the mean intensity across all superpixels within each concentration level was calculated for every multispectral band and plotted as a function of wavelength to obtain the mean spectral signatures shown in Fig.~\ref{fig:spectral_sig}.

\subsection{Linear Regression Model}
\label{subsec:lin_reg}

To establish an interpretable baseline for quantitative urea estimation, a linear regression model was first used to map the spectral information extracted from MSI to the corresponding urea concentration. Let $\mathbf{x}^{T}=[1,\,x_1,\,x_2,\ldots,\,x_{12}]\in\mathbf{R}^{1\times13}$ denote the augmented superpixel feature vector of a milk sample, where $x_1$ to $x_{12}$ represent the mean intensity values of the 12 spectral bands and the first term corresponds to the bias. Let $\mathbf{w}\in\mathbf{R}^{13}$ be the corresponding weight vector. The predicted urea concentration for the $i^{th}$ superpixel sample is then given by $\hat{y}_i=\mathbf{w}^{T}\mathbf{x}^{(i)}$.

Given $N$ training samples, the objective of ordinary linear regression is to minimize the mean squared error (MSE) between the measured and predicted urea concentrations:
\begin{equation}
{MSE} = \frac{1}{N}\sum_{i=1}^{N}\left(y_i-\hat{y}_i\right)^2,
\label{eq:mean_sqloss}
\end{equation}
where $y_i$ and $\hat{y}_i$ denote the measured and predicted urea concentrations of the $i^{th}$ sample, respectively. Although ordinary linear regression provides a direct mapping between spectral features and concentration, it may be sensitive to overfitting when spectral bands are correlated. Therefore, ridge regularization was used to constrain the model complexity by penalizing the squared L2 norm of the weight vector. The resulting regularized cost function is given by
\begin{equation}
J_{\lambda}(\mathbf{w})= \text{MSE} +\frac{\lambda}{2}\left\|\mathbf{w}\right\|^{2}_{2},
\end{equation}
where $\left\|\mathbf{w}\right\|^{2}_{2}$ is the L2 regularization term, and $\lambda \geq 0$ controls the regularization strength.

The design matrix $\mathbf{X}\in \mathbf{R}^{N\times13}$ was formed by stacking the augmented feature vectors of all training samples row-wise, while $\mathbf{y}$ denotes the corresponding vector of measured urea concentrations. The ridge regression solution that minimizes $J_{\lambda}(\mathbf{w})$ was obtained in closed form as
\begin{equation}
\mathbf{w}^{\ast}_{\mathrm{ridge}}=
\left(\mathbf{X}^{T}\mathbf{X}+\lambda \mathbf{I}\right)^{-1}
\mathbf{X}^{T}\mathbf{y}.
\end{equation}
The optimal value of $\lambda$ was selected using grid search combined with $k$-fold cross-validation on the training set.

\subsection{Feed-Forward Neural Network}
\label{subsec:nn}

\begin{figure}[!t]
\centering
\resizebox{\linewidth}{!}{%
\begin{tikzpicture}[
    >=Stealth,
    node distance=2cm and 2.5cm,
    inputNode/.style={circle, draw, fill=blue!30, minimum size=8mm, thick, inner sep=0pt},
    sumNode/.style={circle, draw, fill=orange!40, minimum size=8mm, thick, inner sep=0pt},
    outSumNode/.style={circle, draw, fill=green!50, minimum size=8mm, thick, inner sep=0pt},
    actNode/.style={
        rectangle, draw, fill=gray!20, minimum size=10mm, thick,
        path picture={
            \draw[gray, thin] (0,-0.4) -- (0,0.4); 
            \draw[gray, thin] (-0.4,0) -- (0.4,0);
            \draw[thick] (-0.4,0) -- (0,0) -- (0.35,0.35);
        }
    },
    linearActNode/.style={
        rectangle, draw, fill=gray!20, minimum size=10mm, thick,
        path picture={
            \draw[gray, thin] (0,-0.4) -- (0,0.4); 
            \draw[gray, thin] (-0.4,0) -- (0.4,0);
            \draw[thick] (-0.35,-0.35) -- (0.35,0.35);
        }
    }
]

\node[inputNode] (X1) at (0, 0) {$x_1$};
\node[inputNode] (X2) at (0, -1.5) {$x_2$};
\node at (0, -3.0) {\Large $\vdots$};
\node[inputNode] (X12) at (0, -4.5) {$x_{12}$};

\node[sumNode] (H1S1) at (3, 1.5) {$\Sigma$};
\draw[<-] (H1S1.north) -- ++(0, 0.35) -- ++(0.4, 0) node[right, inner sep=2pt] {$b_1$};
\node[actNode] (H1A1) at (4.5, 1.5) {};
\node[above=0.1cm of H1A1] {A};
\draw[->] (H1S1) -- node[above] {$S_1$} (H1A1);

\node[sumNode] (H1S2) at (3, 0) {$\Sigma$};
\draw[<-] (H1S2.north) -- ++(0, 0.35) -- ++(0.4, 0) node[right, inner sep=2pt] {$b_2$};
\node[actNode] (H1A2) at (4.5, 0) {};
\node[above=0.1cm of H1A2] {A};
\draw[->] (H1S2) -- node[above] {$S_2$} (H1A2);

\node[sumNode] (H1S3) at (3, -1.5) {$\Sigma$};
\draw[<-] (H1S3.north) -- ++(0, 0.35) -- ++(0.4, 0) node[right, inner sep=2pt] {$b_3$};
\node[actNode] (H1A3) at (4.5, -1.5) {};
\node[above=0.1cm of H1A3] {A};
\draw[->] (H1S3) -- node[above] {$S_3$} (H1A3);

\node at (3, -3.0) {\Large $\vdots$};
\node at (4.5, -3.0) {\Large $\vdots$};

\node[sumNode] (H1S63) at (3, -4.5) {$\Sigma$};
\draw[<-] (H1S63.north) -- ++(0, 0.35) -- ++(0.4, 0) node[right, inner sep=2pt] {$b_{63}$};
\node[actNode] (H1A63) at (4.5, -4.5) {};
\node[above=0.1cm of H1A63] {A};
\draw[->] (H1S63) -- node[above] {$S_{63}$} (H1A63);

\node[sumNode] (H1S64) at (3, -6.0) {$\Sigma$};
\draw[<-] (H1S64.north) -- ++(0, 0.35) -- ++(0.4, 0) node[right, inner sep=2pt] {$b_{64}$};
\node[actNode] (H1A64) at (4.5, -6.0) {};
\node[above=0.1cm of H1A64] {A};
\draw[->] (H1S64) -- node[above] {$S_{64}$} (H1A64);

\node[sumNode] (H2S1) at (8, 0.75) {$\Sigma$};
\draw[<-] (H2S1.north) -- ++(0, 0.35) -- ++(0.4, 0) node[right, inner sep=2pt] {$b_1$};
\node[actNode] (H2A1) at (9.5, 0.75) {};
\node[above=0.1cm of H2A1] {A};
\draw[->] (H2S1) -- node[above] {$S_1$} (H2A1);

\node[sumNode] (H2S2) at (8, -0.75) {$\Sigma$};
\draw[<-] (H2S2.north) -- ++(0, 0.35) -- ++(0.4, 0) node[right, inner sep=2pt] {$b_2$};
\node[actNode] (H2A2) at (9.5, -0.75) {};
\node[above=0.1cm of H2A2] {A};
\draw[->] (H2S2) -- node[above] {$S_2$} (H2A2);

\node at (8, -2.25) {\Large $\vdots$};
\node at (9.5, -2.25) {\Large $\vdots$};

\node[sumNode] (H2S31) at (8, -3.75) {$\Sigma$};
\draw[<-] (H2S31.north) -- ++(0, 0.35) -- ++(0.4, 0) node[right, inner sep=2pt] {$b_{31}$};
\node[actNode] (H2A31) at (9.5, -3.75) {};
\node[above=0.1cm of H2A31] {A};
\draw[->] (H2S31) -- node[above] {$S_{31}$} (H2A31);

\node[sumNode] (H2S32) at (8, -5.25) {$\Sigma$};
\draw[<-] (H2S32.north) -- ++(0, 0.35) -- ++(0.4, 0) node[right, inner sep=2pt] {$b_{32}$};
\node[actNode] (H2A32) at (9.5, -5.25) {};
\node[above=0.1cm of H2A32] {A};
\draw[->] (H2S32) -- node[above] {$S_{32}$} (H2A32);

\node[outSumNode] (OutS) at (13, -2.25) {$\Sigma$};
\draw[<-] (OutS.north) -- ++(0, 0.35) -- ++(0.4, 0) node[right, inner sep=2pt] {$b_{out}$};

\node[linearActNode] (OutA) at (14.5, -2.25) {};
\node[above=0.1cm of OutA] {A};
\draw[->] (OutS) -- (OutA);

\draw[->] (OutA) -- ++(1.5, 0) node[right] {$\hat{y}$};

\draw[thick, dashed, draw=gray!80, rounded corners=8pt] (2.2, 2.5) rectangle (5.3, -6.8);
\draw[thick, dashed, draw=gray!80, rounded corners=8pt] (7.2, 1.7) rectangle (10.3, -6.0);


\foreach \i in {1, 2, 12} {
    \foreach \j in {1, 2, 3, 63, 64} {
        \draw[->] (X\i) -- (H1S\j);
    }
}
\node at (1.2, 1.0) {$W_{1,1}$};
\node at (1.2, -5.6) {$W_{12,64}$};

\foreach \i in {1, 2, 3, 63, 64} {
    \foreach \j in {1, 2, 31, 32} {
        \draw[->] (H1A\i) -- (H2S\j);
    }
}

\foreach \i in {1, 2, 31, 32} {
    \draw[->] (H2A\i) -- (OutS);
}
\node at (11.5, -0.6) {$W_{1,out}$};
\node at (11.5, -4.2) {$W_{32,out}$};

\node[font=\bfseries] at (0, -7.5) {INPUT LAYER};
\node[font=\bfseries] at (3.75, -7.5) {HIDDEN LAYER 1};
\node[font=\bfseries] at (8.75, -7.5) {HIDDEN LAYER 2};
\node[font=\bfseries] at (14.5, -7.5) {OUTPUT LAYER};

\end{tikzpicture}
}
\caption{Structure of the feed-forward neural network}
\label{fig:ffnn_architecture}
\end{figure}
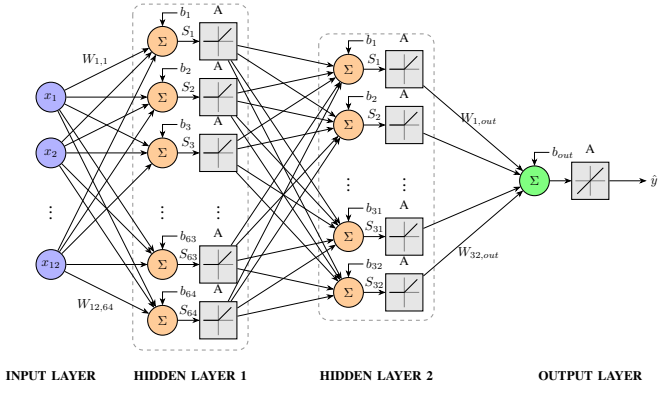

To further analyze the multispectral features using a nonlinear regression framework, a fully connected feed-forward neural network with two hidden layers was employed. As illustrated in Fig.~\ref{fig:ffnn_architecture}, the proposed network consists of an input layer with 12 nodes, two hidden Dense layers with 64 and 32 nodes, respectively, and a single output node. The dataset, composed of the 12 wavelength intensities of each adulterated sample, was z-score normalized and used as input features for training.  

Let $\textbf{x}^{(i)} \in \mathbf{R}^{12}$ represent the input feature vector of the $i^{th}$ sample. The neural network maps the input to the output through successive hidden-layer transformations. For a given hidden layer $l$, an affine transformation is first applied, followed by a nonlinear activation function. In the present study, the hidden layers employed the $relu$ activation function to introduce nonlinearity into the model. Thus, the activation output $\textbf{a}^{[l]}$ for layer $l$ is defined as
\begin{equation}
\textbf{a}^{[l]}=
\max\left(0,\textbf{W}^{[l]T}\textbf{a}^{[l-1]}+\textbf{b}^{[l]}\right)
\end{equation}
where $\textbf{W}^{[l]}$ and $\textbf{b}^{[l]}$ denote the weight matrix and bias vector of layer $l$, respectively, the $\max$ operation is applied element-wise, and the initial input is given by $\textbf{a}^{[0]}=\textbf{x}^{(i)}$.

The output of the second hidden layer is finally mapped by the output layer to obtain the predicted urea concentration. Since the task is formulated as a regression problem, a linear activation is used at the output node. Consequently, the final predicted urea concentration $\hat{y}_i$ is expressed as
\begin{equation}
\hat{y}_i\left(\textbf{x}^{(i)}\right)=\textbf{w}^{[out]T}\textbf{a}^{[2]}+b^{[out]}
\end{equation}
where $\textbf{w}^{[out]} \in \mathbf{R}^{32}$ and $b^{[out]} \in \mathbf{R}$ denote the weight vector and scalar bias of the output layer, respectively.

The MSE loss function, defined in Equation~\eqref{eq:mean_sqloss}, was adopted for training the feed-forward network. Once the loss was computed, backpropagation was performed to compute the gradients of the loss function with respect to each weight. The Adaptive Moment Estimation (Adam) optimizer was utilized with a learning rate of 0.001 to update the weights. The training process employed a batch size of 32 and 150 epochs. In addition, L2 regularization with a coefficient of 0.015 was applied to the network weights to reduce overfitting and improve model generalization.


\section{Results And Discussion}
\label{sec:results}

\begin{figure}[!t]
    \centering
    \includegraphics[width=\columnwidth]{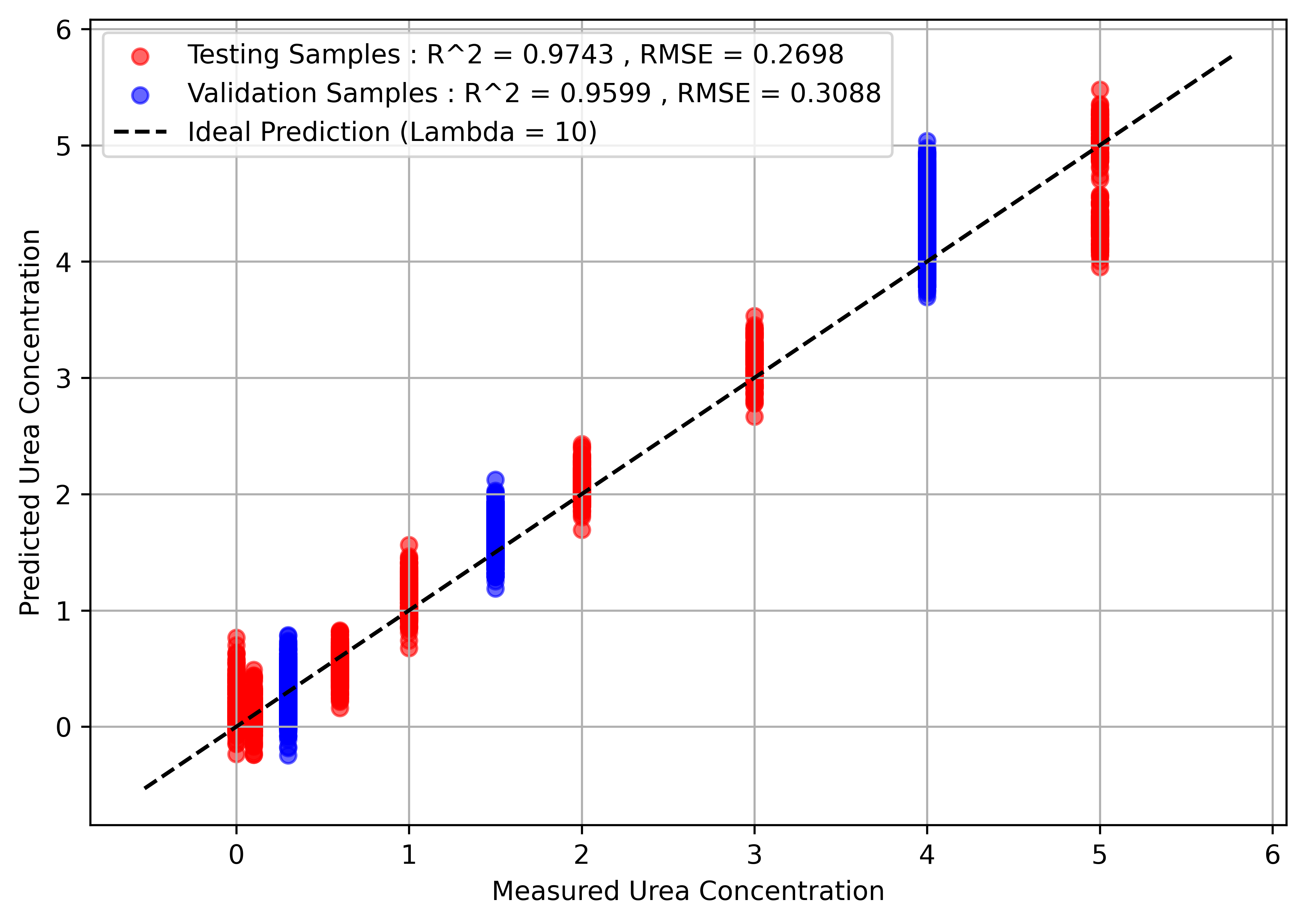}
    \caption{Measured vs predicted values of density balanced urea concentration for regularized linear regression}
    \label{fig:linear_plot}
\end{figure}

\begin{figure}[!t]
    \centering
    \includegraphics[width=\columnwidth]{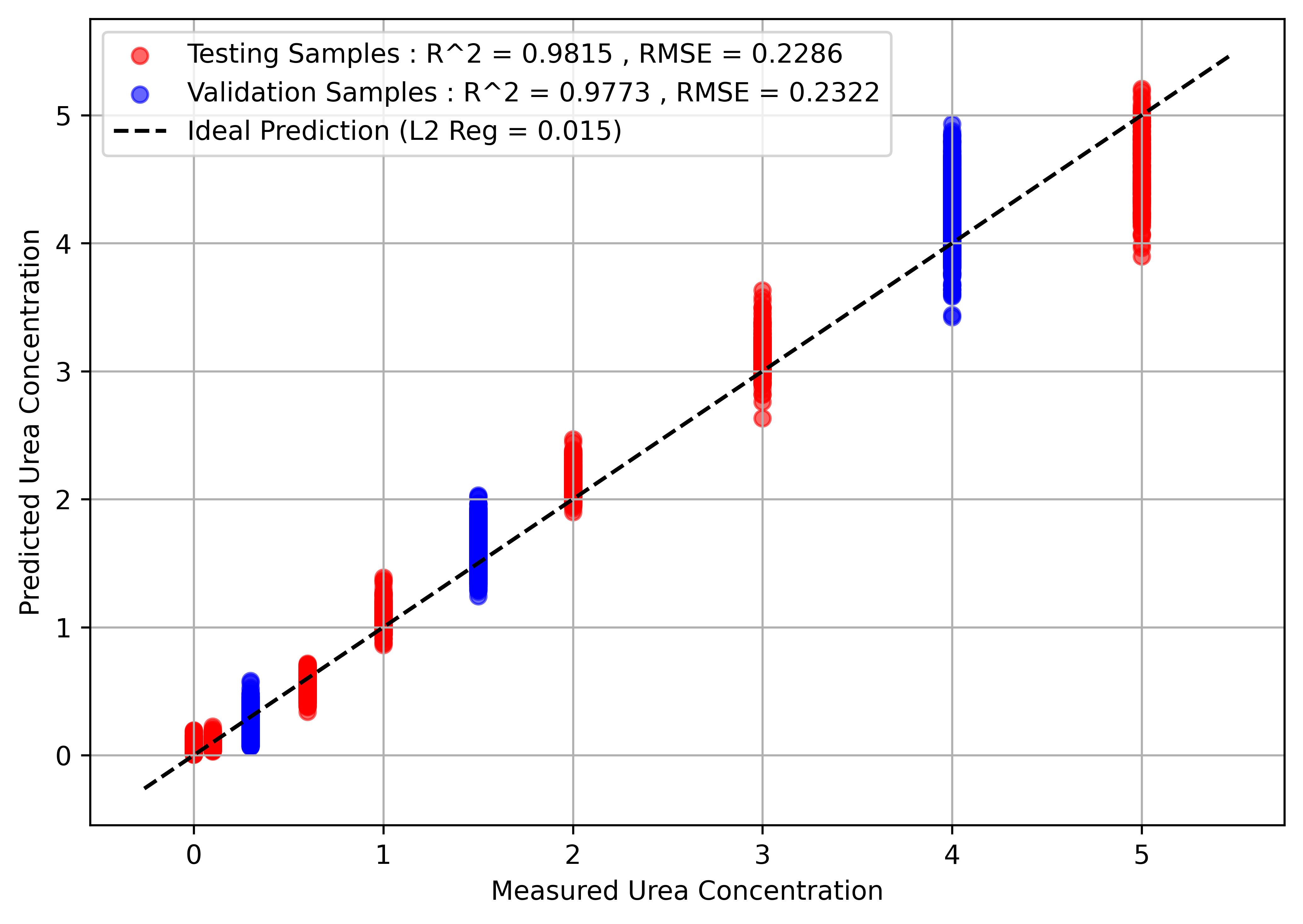}
    \caption{Measured vs predicted values of density balanced urea concentration for feed-forward neural network}
    \label{fig:nn_plot}
\end{figure}

\subsection{Spectral Signature}
\label{subsec:spectral_signature}

The averaged spectral signatures corresponding to different levels or urea adulteration are provided in Figure \ref{fig:spectral_sig}. Upon inspection, we can see that on average, there is an increase in the intensity with increasing concentration, demonstrating that the intensity response of the milk samples varies systematically with different adulteration levels. 

Furthermore distinct differences in the intensity values can be seen for multiple wavelengths, especially concentration-related differences become the most evident at 530 nm, 735 nm, 850 nm, and 940 nm wavelengths. This prominent behavior is confirmed by the wavelength-wise concentration dependence of intensity shown in Fig. \ref{fig:wavelength_vs_adulteration}. However, some deviations from monotonicity can be identified at high concentrations (5.0 g/100 mL). Nevertheless, the results indicate that the proposed transmittance MSI system effectively captures concentration-dependent information suitable for regression modeling.

\subsection{Regression Analysis}
\label{subsec:regression_analysis}

The proposed regression models in Section~\ref{subsec:lin_reg} and Section~\ref{subsec:nn} were used to estimate urea adulteration in milk based on the spectral information extracted from multispectral images.

Initially, quantitative estimation of urea adulteration was carried out using ridge-regularized linear regression. For the training set, the model achieved an \(R^2\) of 0.9798 with an RMSE of 0.2390. For the testing set, the model yielded an \(R^2\) of 0.9743 and an RMSE of 0.2698. Finally, for the validation set, the regression model achieved an \(R^2\) of 0.9599 with an RMSE of 0.3088. Thereafter, a feed-forward neural network was employed to capture the nonlinear relationships within the multispectral data. Compared to the linear regression model, the neural network achieved improved predictive performance across all data splits. For the training set, the model achieved an \(R^2\) of 0.9891 with an RMSE of 0.1753. For the testing set, the corresponding \(R^2\) and RMSE values were 0.9815 and 0.2286, respectively. Finally, for the validation set, the neural network achieved an \(R^2\) of 0.9773 with an RMSE of 0.2322. The reported performance metrics were computed at the independent replicate image-set level by averaging the superpixel predictions within each image set. These results show that the neural network provided better generalization performance than the linear regression model, suggesting that the nonlinear model was more effective in capturing the relationship between the multispectral features and the density-balanced urea concentration.

Overall, both models exhibited strong quantitative predictive performance, confirming the feasibility of the proposed multispectral transmittance framework for urea estimation under controlled density-balanced conditions. The feed-forward neural network provided the strongest performance, accurately predicting the held-out intermediate concentration levels using only 12 spectral bands and demonstrating the potential of the MSI-regression approach for rapid, non-destructive quantitative urea assessment in milk.


\section{Conclusion}
\label{sec:discussion_conclusion}

Rapid and reliable estimation of urea adulteration is essential to ensure the safety and quality of milk. This study demonstrated the feasibility of using transmittance multispectral imaging for non-destructive quantification of urea adulteration in bovine milk under controlled density-balanced laboratory conditions. An in-house-built MSI system capturing 12 spectral bands from 365~nm to 940~nm was used to acquire multispectral responses from milk samples prepared with controlled urea addition and water-based density balancing. The observed concentration-dependent spectral variations confirmed that the acquired MSI data contained useful information for regression-based urea estimation.

Overall, this study establishes that employing transmittance MSI with regression analysis constitutes a powerful, rapid, cost-effective, and non-destructive technique for quantitative urea assessment in milk, with strong potential for practical milk-quality screening. Building on these results, the proposed framework provides a strong basis for dairy supply-chain deployment, with next-stage validation involving naturally variable milk samples from different sources, collection days, and storage conditions, together with benchmarking against other portable screening approaches.



\end{document}